\documentclass[runningheads]{llncs}
\usepackage[T1]{fontenc}
\usepackage{graphicx,verbatim}
\usepackage{amsmath,amssymb}
\usepackage{booktabs}
\usepackage{amsfonts}
\usepackage{xcolor}
\usepackage{algorithm}
\usepackage{algpseudocode}
\usepackage{marvosym}  

\begin{document}
\title{IMVS: Interactive Medical Volume Segmentation with Test-Time Adaptation - A New Method for Annotating Radiology Datasets\thanks{Accepted at the HAIC workshop, MICCAI 2026. This is the Submitted Manuscript, prior to peer review; it has not undergone peer review or any post-submission improvements or corrections. The Version of Record will be published in \emph{Medical Image Computing and Computer Assisted Intervention -- MICCAI 2026 Satellite Events}, Lecture Notes in Computer Science, Springer, and a DOI link will be added here once it is available.}}
%
\titlerunning{IMVS: Interactive Medical Volume Segmentation}

\author{Abhilaksh Singh~Reen\thanks{Equal contribution.}\fnmsep\inst{1}$^{(\textrm{\Letter})}$ \and
Kushal~Borkar$^{\star\star}$\fnmsep\inst{2} \and
Ritvik~Mahapatra$^{\star\star}$\fnmsep\inst{3}}
\authorrunning{A. S. Reen et al.}
\institute{Independent Researcher, Delhi, India\\
    \email{abhilakshsinghreen@gmail.com}
\and
    Independent Researcher, Kolkata, India\\
    \email{kushalborkar.11@gmail.com}
\and
    California State University, Fresno, CA, USA\\
    \email{ritvik123@mail.fresnostate.edu}}

\maketitle              
\begin{abstract}
Constructing large annotated radiology datasets is bottlenecked by the manual effort of delineating structures slice-by-slice in 3D volumes. Interactive methods reduce this effort but stay interaction-inefficient: slice-wise methods (including many foundation models) ignore inter-slice continuity, while 3D and video-based methods propagate a prompt with a \emph{fixed} propagator that never adapts to the target volume, so it drifts on low-contrast or pathological structures and must be re-prompted. We present IMVS, a human-in-the-loop annotation framework that, rather than a new segmentation primitive, composes three components into a closed loop: a lightweight 2D Slice Mask Adapter (SMA) fine-tuned online from user scribbles, a frozen Volume Mask Tracker (VMT) that propagates corrected masks across adjacent slices, and a soft teacher--student alignment that limits forgetting. The SMA is backbone-agnostic (UNet++, DeepLabV3, TransUNet). Across 8 public CT/MRI datasets, IMVS matches strong interactive baselines in quality while substantially cutting annotation effort: $14.4\times$ faster than a proficient copy-based manual workflow ($22.3\times$ over naive manual), $4.6\times$ over slice-wise and $1.9\times$ over 3D interactive methods. Foundation models such as MedSAM2 and ScribblePrompt stay competitive or stronger on well-delineated organs; IMVS's advantage is largest on challenging targets and on overall interaction efficiency. Source code and Demo Video: https://github.com/AbhilakshSinghReen/imvs.

\end{abstract}

\section{Introduction}

Interactive segmentation frameworks, including foundation models such as SAM~\cite{segmentanything2023}, MedSAM~\cite{medsegmentanything2024}, and ScribblePrompt~\cite{wong2024scribbleprompt}, reduce annotation effort by taking human prompts (boxes~\cite{xu2017deep}, clicks~\cite{fbrs2020,ritm2022,isegformer2022}, or scribbles~\cite{wu2014milcut,agustsson2019interactive}) directly into the inference loop. Applied to volumes, slice-wise methods refine individual masks well but ignore inter-slice continuity, producing redundant interactions across a volume. Recent 3D and video-based interactive methods: PRISM~\cite{li2024prism}, nnInteractive~\cite{isensee2025nninteractive}, and video-propagation systems such as SAM2/MedSAM2, instead, carry a prompt through the volume, but rely on a \emph{fixed}, pre-trained propagator that is never adapted to the volume being annotated. Under distribution shift, low contrast, and pathology typical of clinical targets, such static propagators drift and must be re-prompted, so effort is still spent correcting the same kinds of errors repeatedly.

Motivated by the temporal consistency exploited in Video Object Segmentation (VOS), we build a human-in-the-loop framework, IMVS, that likewise propagates masks across adjacent slices but unlike the fixed 3D propagators above, IMVS couples propagation with \emph{online adaptation}: each correction fixes the current slice \emph{and} updates a lightweight per-slice model for the slices ahead, so an interaction is reused over many slices instead of being repeated on each one, and the system stops repeating errors rather than merely re-propagating them. In the taxonomy of interactive medical segmentation~\cite{marinov2024deep}, IMVS is scribble-driven and iteratively refined, distinguished by \emph{where} human effort is spent: it amortizes corrections over a volume through learned propagation and online adaptation. We therefore position it not as a new segmentation primitive but as a \emph{human-AI collaboration system} for annotation efficiency, whose building blocks: interactive 2D segmentation, teacher-student test-time adaptation~\cite{wang2020tent}, soft alignment against forgetting~\cite{kirkpatrick2017overcoming}, and attention-based propagation - are adapted from established paradigms; the contribution is how they compose into an efficient closed loop.

Concretely, we provide: (1) a closed-loop pipeline that couples a lightweight, online-adapted 2D Slice Mask Adapter (SMA) with a frozen Volume Mask Tracker (VMT), confining backpropagation to a small network while a fixed propagator enforces volumetric consistency; (2) a soft teacher-student alignment enabling continual refinement from user scribbles without catastrophic forgetting across slices and volumes; and (3) an empirical study on 8 CT/MRI datasets with a 9-annotator user study, quantifying annotation-efficiency gains and characterizing where the approach helps and where it does not. The SMA is backbone-agnostic (validated on UNet++, DeepLabV3, TransUNet).

\section{Related Work}

Interactive methods refine masks from user inputs: boxes~\cite{xu2017deep}, clicks~\cite{fbrs2020,ritm2022,isegformer2022,pseudoclick2022,focalclick2022}, or scribbles~\cite{wu2014milcut,agustsson2019interactive}; f-BRS~\cite{fbrs2020} and RiTM~\cite{ritm2022} introduced iterative inference-time refinement, and iSegFormer~\cite{isegformer2022} a transformer backbone for medical images. Foundation models (SAM~\cite{segmentanything2023}, MedSAM~\cite{medsegmentanything2024,ma2024segment}) generalize across targets but degrade under medical distribution shift and are typically applied per-slice; lighter 3D approaches such as PRISM~\cite{li2024prism} and promptable systems like nnInteractive~\cite{isensee2025nninteractive} address volumes more directly. We refer to Marinov et al.~\cite{marinov2024deep} for a systematic taxonomy; within it IMVS is scribble-driven and iteratively-refined, distinguished by amortizing interactions across slices rather than maximizing single-mask accuracy~\cite{scribbleprompt2024,lin2016scribblesup}.

Test-time and continual adaptation avoid target-domain retraining using uncertainty~\cite{wang2020tent,liang2020we} or teacher pseudo-labels~\cite{wang2022continual}; mean-teacher formulations are effective for continual TTA~\cite{dobler2023robust} but are \emph{unsupervised}. IMVS differs by being \emph{interactive and weakly supervised}: the dominant signal is the user's corrective scribble, with teacher/propagation terms only stabilizing, which counteracts the pseudo-label drift to which unsupervised teacher-student adaptation is prone~\cite{wang2022continual}, while soft alignment guards against forgetting~\cite{kirkpatrick2017overcoming,french1999catastrophic}. Propagating an annotation to neighbouring slices is long-standing---Wang et al.~\cite{wang2016online} used online random forests for placental segmentation, and VOS trackers later formalized memory/attention-based propagation. IMVS shares this intuition and, like modern VOS trackers, uses a learned attention-based propagator with a memory bank: pretrained on video object segmentation~\cite{xu2018youtube} and fine-tuned for medical propagation. What distinguishes IMVS is that this propagator is deliberately kept lightweight and frozen during the annotation loop, with its errors recovered by online SMA adaptation rather than by a heavier tracker. This division of labour - a fixed propagator plus an adaptive per-slice model, is the main design choice.

\section{Method}
\label{sec:method}

\begin{figure}[t]
  \centering
  \includegraphics[width=\textwidth]{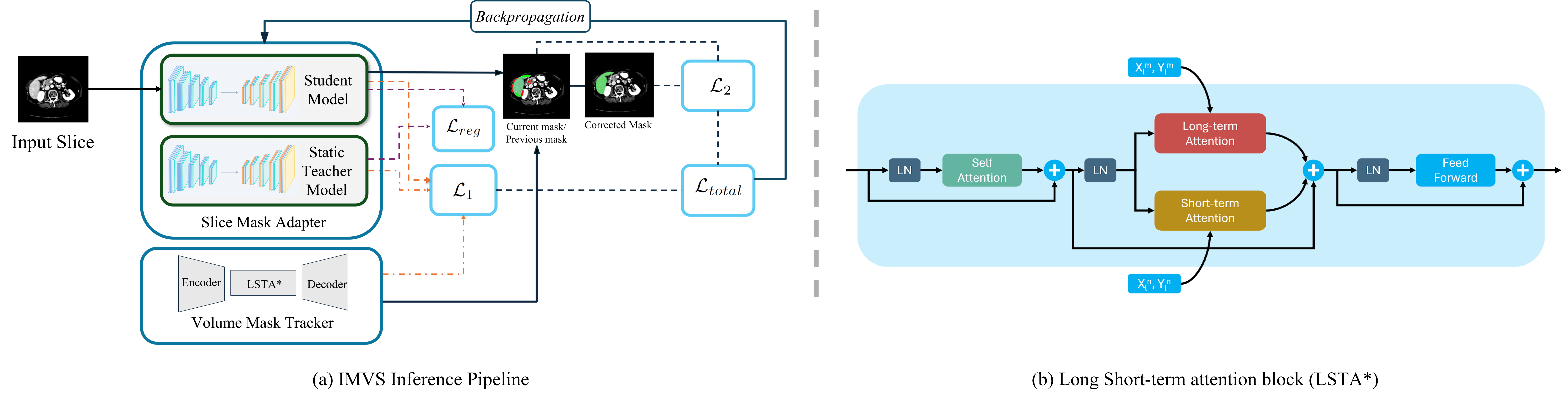}
  \caption{Overview of the IMVS closed annotation loop. The SMA predicts a mask the user refines with scribbles (teacher--student online adaptation); the frozen VMT propagates the corrected mask across adjacent slices via long/short-term attention and a memory bank.}
  \label{fig:pipeline}
\end{figure}

\textbf{Problem setting and loop.} IMVS couples a lightweight 2D SMA, fine-tuned online from scribbles, with a frozen VMT propagator, confining backpropagation to the small SMA while the fixed propagator enforces volumetric consistency. Let $f_{\theta_0}$ be a model pre-trained (parameters $\theta_0$) on source data $D_s$. At step $t$ the model sees a slice $x_t\in\mathbb{R}^{H\times W}$, the previous slice/mask $(x_{t-1},\hat{y}_{t-1})$, and (on intervention) scribbles $\mathcal{S}_t$, and the user refines the prediction into an accepted mask $m_t$; the goal is to segment an unseen volume with as few interactions as possible without forgetting $D_s$. The annotator begins at the slice $x_0$ where the target first becomes clearly visible - a \emph{protocol} choice (mirroring how experts scroll to a structure's onset), not a tuned hyperparameter. Because the loop re-seeds from \emph{any} corrected slice, a different start costs at most one early correction, not final quality. From $x_0$ the framework alternates an \emph{Interactive Mode} (scribbles drive online SMA adaptation) and an \emph{Automatic Mode} (the VMT propagates the accepted mask forward for up to $K$ slices. We use $K{=}17$, trading propagation error against correction frequency. Before a correction is required; $K$ is justified empirically in Sec.~\ref{sec:component_ablation}). Algorithm~\ref{alg:imvs} summarizes the loop for one volume: the user intervenes only when a propagated mask is unsatisfactory, and each correction fixes the current slice and adapts the student for the slices ahead.

\subsection{Slice Mask Adapter (SMA)}
\label{sec:sma}
The SMA is a standard, interchangeable 2D segmentation network; we evaluate UNet++, DeepLabV3+, and TransUNet, and adopt UNet++ as default. It is instantiated as a teacher--student pair: a frozen teacher $f_{\theta^T}$ and an adaptive student $f_{\theta^S}$, both initialized from the same weights $\theta^T=\theta^S=\theta_0$. The teacher preserves source-domain knowledge; only the student is updated online. The teacher and the VMT are \emph{never} updated at test time.

\textbf{How scribbles enter (early fusion).} Foreground/background scribbles $\mathcal{S}_t=\{s^{fg}_t,s^{bg}_t\}$ are rasterized (dilation radius 3\,px) into two binary guidance channels $M_{fg},M_{bg}\in\{0,1\}^{H\times W}$, set to $1$ on foreground/background strokes respectively, and \emph{concatenated with the grayscale slice at the input}, giving a 3-channel input (CT/MRI $+\,M_{fg}+M_{bg}$) to the SMA. Interactions are thus fused early, keeping them backbone-agnostic.

\subsection{Volume Mask Tracker (VMT)}
\label{sec:vmt}
The VMT maps $(x_t, m_{t-1}, \mathcal{M})\mapsto \hat{y}^{VMT}_t$, propagating the accepted mask of slice $t{-}1$ onto slice $t$ using a memory bank $\mathcal{M}$ of recent (feature, mask) pairs. Its encoder is a \textbf{ViT-B/16} pretrained on video object segmentation (YouTube-VOS 2019~\cite{xu2018youtube}) and fine-tuned on medical volumes (Sec.~\ref{sec:training}), producing
\begin{equation}
    F_t^{enc} = \text{Encoder}(x_t) \in \mathbb{R}^{H/16 \times W/16 \times D},\qquad D=768.
\end{equation}
The previous mask is encoded to the same resolution (a $1{\times}1$ convolution on the resized mask) and concatenated with slice features to inject temporal, object-specific context. A transformer with $L=6$ Long+Short-Term Attention (LSTA) layers applies scaled dot-product attention in two scopes: \emph{short-term} over the two preceding slices (frame-to-frame smoothness) and \emph{long-term} over a memory buffer of the six most recent slices (appearance consistency and drift prevention). A decoder upsamples propagated mask $\hat{y}^{VMT}_t$. VMT is frozen at inference.

\subsection{Online Adaptation Objective}
\label{sec:objective}
We call this \emph{test-time adaptation} to indicate \emph{when} the SMA is updated (at inference, on the target volume), not unsupervised adaptation~\cite{wang2020tent,dobler2023robust}: the dominant signal is the user's corrective scribble, with propagated/teacher pseudo-labels only for stabilization. At each interactive step the student minimizes three terms. (A) A \emph{consistency} term distills the frozen VMT mask and the augmentation-averaged teacher prediction $\hat{y}^{aug}_t$ into the student $\hat{y}^S_t$,
\begin{equation}
    \mathcal{L}_1 = -\!\sum_{c} \big[\, \beta_1\, \hat{y}^{VMT}_t[:,c]\log(\hat{y}^S_t[:,c]) + \beta_2\, \hat{y}^{aug}_t[:,c]\log(\hat{y}^S_t[:,c]) \big],
\end{equation}
with class index $c$ and weights $\beta_1,\beta_2$ biased toward VMT guidance. (B) An \emph{interactive} cross-entropy term $\mathcal{L}_2=-\sum_c \hat{y}^{corrected}_t[:,c]\log(\hat{y}^S_t[:,c])$ matches the user-corrected mask $\hat{y}^{corrected}_t$, and (C) a soft \emph{alignment} term prevents forgetting by anchoring the student's batch-norm (BN) parameters to the teacher, $\mathcal{L}_{reg}(\theta_t)=\sum_l \mathbb{1}[l\in\text{BN}]\,\|\theta^S_l-\theta^T_l\|_2^2$. The total objective is
\begin{equation}
\label{eq:total_loss}
    \mathcal{L}_{total} = \mathcal{L}_1 + \mathcal{L}_2 + \kappa\,\mathcal{L}_{reg}.
\end{equation}
The student is updated by SGD only when $\mathcal{L}_{total}>\gamma=0.05$ (suppressing noisy updates), with $\kappa=0.1$. Freezing the propagator prevents pseudo-label drift from compounding; the adapted student is carried to subsequent volumes as an efficiency warm-start (supervision stays user-driven, no bias to annotation).

\begin{algorithm}[t]
\small
\caption{IMVS interaction loop for one volume}
\label{alg:imvs}
\begin{algorithmic}[1]
\Require volume $\{x_i\}$; SMA $f_{\theta_0}$; frozen VMT $g$; threshold $\gamma$; span $K$
\State $\theta^T,\theta^S \gets \theta_0$; user seeds $x_0$ with $\mathcal{S}_0$ (Interactive Mode); refine and update $\theta^S$; $m_0\gets\hat{y}^{corrected}_0$; push to $\mathcal{M}$
\For{$t=1,2,\dots$ until the volume is covered}
  \State $\hat{y}^{VMT}_t \gets g(x_t, m_{t-1}, \mathcal{M})$ \Comment{automatic propagation}
  \If{user accepts $\hat{y}^{VMT}_t$ \textbf{and} span since correction $<K$}
     \State $m_t \gets \hat{y}^{VMT}_t$
  \Else\; \Comment{correct and adapt}
     \State user draws $\mathcal{S}_t$; if $\mathcal{L}_{total}>\gamma$ update $\theta^S$ (Eq.~\ref{eq:total_loss}); $m_t \gets \hat{y}^{corrected}_t$
  \EndIf
  \State push $(F_t,m_t)$ to $\mathcal{M}$, evict oldest
\EndFor
\State carry $\theta^S$ to the next volume \Comment{warm-start}
\end{algorithmic}
\end{algorithm}

\subsection{Human-AI Interaction Workflow}
\label{sec:workflow}
IMVS is designed around iterative human-AI collaboration rather than one-shot segmentation. The annotator intervenes only when propagated masks become unacceptable; each correction is immediately incorporated by the adaptive SMA and reused across subsequent slices through propagation, a capability missing in existing methods. Thus, user input serves not only as local error correction but also as guidance that adapts future model behavior, reducing repeated corrections within the same session. This shifts the annotator's role from refining every slice to supervising and steering an adaptive segmentation assistant.

\subsection{Training and Implementation}
\label{sec:training}
\textbf{Data / zero-shot split.} Both networks are trained \emph{only} on stratified 70:15:15 splits of three datasets (BraTS, LiTS, MSD Pancreas; the ``consolidated'' set); the other five (CHAOS-CT/MRI, AMOS-CT, MSD Prostate, MSD HepaticVessel) are never seen in training and evaluated \textbf{zero-shot}, so cross-dataset results measure generalization. \textbf{SMA:} ImageNet-initialized backbones, slices $256\times256$, 110 epochs (AdamW, lr $1\!\times\!10^{-4}$, batch 16), loss $\mathcal{L}_{Dice}+\mathcal{L}_{CE}$; training scribbles simulated from ground truth (boundary default). Online adaptation uses SGD (momentum 0.9, lr $1\!\times\!10^{-4}$) with horizontal-flip and intensity-jitter test-time augmentation on the teacher. \textbf{VMT:} the ViT-B propagator is pretrained on YouTube-VOS 2019~\cite{xu2018youtube} (natural-video object segmentation) and then fine-tuned as a mask propagator on the consolidated set for 58 epochs (Adam, lr $2\!\times\!10^{-4}$, batch 8) with a CE+Dice loss between propagated mask and ground truth, then frozen at inference. \textbf{Hardware:} a single NVIDIA RTX 5090 (CUDA 12.8).

Training uses \emph{simulated} scribbles. For training only, metrics use a deterministic simulator that reacts to the current error region until the stopping criterion is met. This is reproducible and annotator-independent.

\section{Experiments and Results}
\label{sec:experiments}

\begin{table}[t]
\centering
\caption{Annotation efficiency per method and dataset, shown as \textbf{time} (minutes) with \textbf{\# interactions} in parentheses. Manual baselines have no discrete interactions ($-$). \textbf{Manual With Copy} is a proficient copy-and-edit workflow (the realistic manual reference). Best (lowest) time and interactions per column in bold.}
\label{tab:seg_time}
\scriptsize
\setlength{\tabcolsep}{4pt}
\resizebox{\textwidth}{!}{%
\begin{tabular}{lcccccccc}
\toprule
\textbf{Method} & \textbf{CHAOS CT} & \textbf{CHAOS MRI} & \textbf{AMOS CT} & \textbf{MSD Prostate} & \textbf{LiTS} & \textbf{MSD Hep.Ves.} & \textbf{MSD Pancreas} & \textbf{BraTS} \\
\midrule
\textbf{Manual}          & 44.79 ($-$) & 21.5 ($-$) & 65.5 ($-$) & 8.53 ($-$) & 73.53 ($-$) & 96.3 ($-$) & 8.11 ($-$) & 55.26 ($-$) \\
\textbf{Manual w/ Copy}  & 28.83 ($-$) & 13.68 ($-$) & 42.79 ($-$) & 5.4 ($-$) & 47.56 ($-$) & 61.85 ($-$) & 5.26 ($-$) & 35.52 ($-$) \\
\textbf{f-BRS}           & 7.23 (10) & 3.41 (9) & 10.55 (34) & 1.64 (16) & 17.57 (27) & 22.8 (126) & 1.59 (23) & 11.49 (30) \\
\textbf{Med SAM}         & 12.61 (13) & 6.02 (11) & 18.51 (31) & 2.39 (18) & 20.51 (28) & 28.25 (122) & 2.27 (22) & 15.55 (30) \\
\textbf{ScribblePrompt}  & 7.79 (20) & 3.65 (12) & 11.36 (29) & 1.45 (21) & 12.68 (25) & 18.4 (121) & 1.39 (24) & 9.41 (29) \\
\textbf{iSegFormer}      & 7.12 (8) & 3.36 (10) & 10.42 (10) & 1.35 (6) & 11.58 (10) & 15.4 (32) & 1.32 (8) & 8.73 (9) \\
\textbf{PRISM}           & 5.49 (6) & 2.65 (4) & 8.06 (9) & 1.07 (3) & 3.93 (5) & 14.7 (28) & 1.03 (3) & 6.79 (8) \\
\textbf{MedSAM2}         & \textbf{1.89} (\textbf{3}) & \textbf{0.96} (\textbf{2}) & \textbf{2.74} (\textbf{3}) & 0.93 (3) & 4.17 (5) & 11.3 (\textbf{19}) & 0.76 (3) & 3.1 (4) \\
\textbf{nnInteractive}   & 1.94 (4) & 0.965 (3) & 2.81 (4) & 0.7 (3) & 3.705 (\textbf{4}) & 11 (23) & 0.61 (4) & 2.75 (5) \\
\textbf{Ours (UNet++)}   & 1.99 (4) & 0.97 (3) & 2.88 (4) & \textbf{0.47} (\textbf{2}) & \textbf{3.24} (\textbf{4}) & \textbf{10.7} (20) & \textbf{0.46} (\textbf{2}) & \textbf{2.4} (\textbf{3}) \\
\bottomrule
\end{tabular}%
}
\vspace{-3mm}
\end{table}

We evaluate across \textbf{8 CT/MRI benchmark datasets}, focusing on challenging cases (liver tumors, pancreatic neoplasms, hepatic vasculature); training uses only the BraTS/LiTS/MSD-Pancreas splits, and the other five datasets are held out zero-shot. We benchmark seven interactive methods: MedSAM, MedSAM2, ScribblePrompt, PRISM, f-BRS, iSegFormer, and nnInteractive~\cite{isensee2025nninteractive}. Interaction modality is heterogeneous (MedSAM/MedSAM2 and PRISM use boxes/clicks; ScribblePrompt, f-BRS, iSegFormer, and IMVS use scribbles/clicks; nnInteractive supports all three). We account for this when attributing the efficiency gains.

\subsection{User Study and Stopping Criterion}
\label{sec:user_study}
Annotation times were measured in a user study with \textbf{nine residents}, reflecting a realistic annotator population for radiology dataset construction under expert supervision. Participants had varying experience and evaluated all methods using same acceptance criterion: advancing once the segmentation was qualitatively satisfactory. Aggregated accepted masks defined the automatic stopping threshold ($\mathrm{DSC}=0.885$, $\mathrm{NSD}=0.894$, $\mathrm{HD95}=4\,\mathrm{mm}$). Inter-annotator agreement was high (mean pairwise $\mathrm{DSC}=0.89\pm0.04$), with consistent acceptance thresholds across raters ($\mathrm{DSC}=0.885\pm0.03$), supporting its use as a shared comparison rule. The outcome of the study (Table~\ref{tab:seg_time}) is human workload reduction - both in time as well as number of interactions.

\subsection{Annotation Efficiency}
\label{sec:efficiency}
\textbf{Time and interactions.} Table~\ref{tab:seg_time} reports both annotation time and interaction counts. IMVS is fastest on 5 of 8 datasets and fastest in aggregate, but not uniformly best. MedSAM2 is faster on the well-delineated healthy organs of CHAOS-CT/MRI and AMOS-CT, this is reported in the per-dataset study rather than being averaged away. On interaction count, IMVS is most efficient on 4 of 8 datasets (2 on MSD Prostate/Pancreas, 3 on BraTS, 4 on LiTS) and stays within one interaction of MedSAM2 elsewhere. The tortuous hepatic vessels are the hard case for \emph{every} method: IMVS still needs 20 interactions there (vs.\ 19 for MedSAM2 and 28-126 for the slice-wise baselines), consistent with our claim that the approach helps least on discontinuous structures.

\textbf{Sources of the gain.}
\label{sec:efficiency_sources}
We state speedups conservatively: IMVS is $22.3\times$ faster than naive \textbf{Manual} annotation but $\mathbf{14.4\times}$ faster than the fairer \textbf{Manual With Copy} (copy-and-edit the previous mask). Part of the per-interaction gain over click/box baselines is modality (a scribble carries more information than a click); comparing against \emph{scribble} baselines under the identical criterion isolates the algorithmic contribution: amortizing interactions over a propagated window plus online SMA adaptation. IMVS also uses the least compute ($3.23$\,GB VRAM, $15.93$\,s GPU/volume). A paired Wilcoxon signed-rank test (per volume, MedSAM2) locates the real advantage: significant over the four challenging datasets (LiTS, MSD Pancreas, MSD HepaticVessel, BraTS; $p=0.02$) but similar over the well-delineated ones (CHAOS-CT/MRI, AMOS, MSD Prostate; $p=0.26$).

\subsection{Segmentation Quality}
\begin{table}[t]
\centering
\caption{Interaction efficiency at matched quality: number of user interactions needed to reach DSC $\geq 0.90$ (near the stopping operating point) per dataset; lower is better. This complements Table~\ref{tab:seg_time} (effort to satisfy the stopping criterion). IMVS reaches the target with the fewest interactions on every dataset.}
\label{tab:per_dataset_metrics}
\scriptsize
\setlength{\tabcolsep}{4pt}
\resizebox{\textwidth}{!}{%
\begin{tabular}{lcccccccc}
\toprule
\textbf{Method} & \textbf{CHAOS-CT} & \textbf{CHAOS-MRI} & \textbf{AMOS} & \textbf{Prostate} & \textbf{LiTS} & \textbf{Pancreas} & \textbf{BraTS} & \textbf{Mean} \\
\midrule
iSegFormer     & 13 & 14 & 13 & 12 & 11 & 15 & 13.5 & 13.1 \\
PRISM          & 12 & 13 & 12 & 11 & 10 & 14 & 12.5 & 12.1 \\
MedSAM2        & 10 & 11 & 10 & 9  & 8  & 12 & 10.5 & 10.1 \\
\textbf{Ours}  & \textbf{7} & \textbf{8} & \textbf{7} & \textbf{6} & \textbf{6} & \textbf{8} & \textbf{7.5} & \textbf{7.1} \\
\bottomrule
\end{tabular}%
}
\vspace{-3mm}
\end{table}

Table~\ref{tab:per_dataset_metrics} complements Table~\ref{tab:seg_time}: per dataset, how many interactions each method needs to \emph{reach} a demanding target (DSC $\geq 0.90$, near the operating point). IMVS needs the fewest on every dataset: $7.1$ on average vs.\ $10.1$ for MedSAM2 and $13.1$ for slice-wise iSegFormer ($1.4$--$1.8\times$ fewer). Final quality is closely clustered once enough interactions are spent, so IMVS's contribution is reaching it \emph{sooner}, not exceeding the plateau. IMVS thus matches strong baselines in accuracy while being far more interaction and VRAM-efficient; it is not claimed best in raw quality on every dataset. Robustness: On tortuous hepatic vessels the VMT tracks reliably for only $\sim$4--6 slices (vs.\ up to 17 on well-defined organs), so more corrections are needed. Every method degrades here. IMVS stays competitive because online SMA adaptation recovers tracking with typically a single scribble.

\subsection{Ablation: Contribution of Each Component}
\label{sec:component_ablation}
Table~\ref{tab:where_to_adapt} isolates the components on the consolidated set. \textbf{Online SMA adaptation} is the largest factor: without it (\textit{No SMA, VMT frozen}) the loop needs 18 interactions at DSC $0.86$; adding it (our method) cuts interactions to 4 at DSC $0.92$ and raises the tracked span from 3 to 12 slices. \textbf{Freezing the VMT} is essential: adapting it as well gives no accuracy gain but $6\times$ the VRAM, $14\times$ the latency, and a shorter tracked span (caused by temporal drift). Among backbones, UNet++ beats DeepLabV3/TransUNet across all metrics (dense skip connections aid boundary refinement under sparse scribbles) and is the default; a memory-bank size of 6 is optimal (4--5 degrade tracking, 7--9 add compute without gains). The propagation length $K{=}17$ is likewise chosen from a sweep $K\in\{5,\dots,25\}$: smaller $K$ forces frequent corrections (7.2 interactions at $K{=}5$ vs.\ 4.1 at $K{=}17$), while larger $K$ raises the propagation failure rate (29\% at $K{=}17$ to 55\% at $K{=}25$) as masks drift.

\begin{table}[t]
\centering
\caption{Ablation of adaptation strategies on the consolidated set (per-volume averages). \textit{SMA (adaptive), VMT frozen} is our method; \textit{Tracked Slices} is the average span before a correction is needed.}
\label{tab:where_to_adapt}
\resizebox{\textwidth}{!}{%
\begin{tabular}{lccccccc}
\toprule
\textbf{Approach} & \textbf{\# Int.} & \textbf{\# Tracked} & \textbf{Dice} & \textbf{NSD} & \textbf{HD95} & \textbf{Max VRAM} & \textbf{GPU Time} \\
\midrule
No SMA, VMT frozen & 18 & 3 & 0.86 & 0.87 & 2.7 & 1.7 GB & 4ms \\
Ours (Memory Size 6) & 4 & 12 & 0.92 & 0.91 & 1.6 & 3.23 GB & 129ms \\
SMA (adaptive), VMT adapted & 8 & 7 & 0.92 & 0.91 & 1.6 & 20.7 GB & 1857ms \\
\bottomrule
\end{tabular}%
}
\vspace{-3mm}
\end{table}

\section{Scope and Limitations}
\label{sec:scope}
IMVS segments a \emph{single} target per pass; multi-label volumes use sequential single-label sessions sharing one adaptation context, and native multi-object propagation is future work. Efficiency comes from amortizing interactions over a propagated span, so IMVS helps least on tortuous or discontinuous structures, where more interactions are needed.

\section{Conclusion}
We have presented IMVS: a human-in-the-loop framework that composes an online-adapted 2D SMA, a frozen VMT propagator, and soft teacher-student alignment into a closed annotation loop. Its benefit is annotation efficiency: propagating each correction across slices and adapting the SMA online cuts interactions, time, and VRAM. We view IMVS as a new method in human-AI collaboration for efficient dataset construction. Future extensions include topology-aware and multi-object propagation.

%
%
%
\bibliographystyle{splncs04}
\bibliography{references}

\end{document}